\documentclass{article}
\usepackage{spconf,amsmath,graphicx}
\usepackage{subcaption}
\usepackage{float}
\usepackage[hyphens]{url}
\usepackage{booktabs}

\usepackage{enumitem}
\setlist{nosep, leftmargin=14pt}

\usepackage{mwe} 

\title{Towards 1000-fold Electron Microscopy Image Compression for Connectomics via VQ-VAE with Transformer Prior}

\name{Fuming Yang \quad Yicong Li \quad Hanspeter Pfister \quad Jeff W. Lichtman \quad Yaron Meirovitch}
\address{Harvard University\\
\small \textit{Corresponding authors:} \texttt{fumingyang@fas.harvard.edu},
\texttt{jeff@mcb.harvard.edu},
\texttt{yaron.mr@gmail.com}}

\begin{document}
%
\maketitle
\begin{abstract}
\noindent Petascale electron microscopy (EM) datasets push storage, transfer, and downstream analysis toward their current limits. We present a vector-quantized variational autoencoder-based (VQ-VAE) compression framework for EM that spans $16{\times}$ to $1024{\times}$ and enables \emph{pay-as-you-decode} usage: top-only decoding for extreme compression, with an optional Transformer prior that predicts bottom tokens (without changing the compression ratio) to restore texture via feature-wise linear modulation (FiLM) and concatenation; we further introduce an ROI-driven workflow that performs selective high-resolution reconstruction from $1024{\times}$-compressed latents only where needed.
\end{abstract}

\section{Introduction}
\noindent EM connectomics has seen orders-of-magnitude growth in data volume: from early GB-scale datasets (e.g., the complete nervous system of \emph{C.\ elegans} \cite{White1986Celegans} to the TB-scale adult fruit fly brain \cite{Scheffer2020Fly}), and now to PB-scale cubic millimeter volumes of human \cite{H01} and mouse \cite{MICrONSConsortium2025Nature} cortex. At present, the High-throughput Integrative Mouse Connectomics (Hi\mbox{-}MC) effort is imaging an entire mouse hippocampus (about 20\,PB) and is moving toward whole mouse brain dataset that is approaching EB. These scales strain storage and inter-site transfer, as well as downstream 3D reconstruction and computational analysis.

To address these challenges, we propose a compression framework based on a vector-quantized variational autoencoder \cite{Oord2017VQVAE} for large-scale EM datasets. The framework has two aims: (i) achieving extreme compression while preserving segmentation accuracy, thereby reducing compute and reconstruction time; and (ii) attaining the highest feasible compression ratio while preserving neuronal structures. Compared with the strong baseline AVIF \cite{AVIF}, we report head-to-head results at $16{\times}$ and $64{\times}$ on two representative datasets (H01-human \cite{H01} and mouse cerebellum \cite{Dhanyasi2025DevCerebellum}), showing nearly identical, stable performance. Relative to prior EM-oriented VAE-based compression \cite{Li2025EMCompressor}, our approach, to the best of our knowledge \cite{minnen2021denoising,Li2025EMCompressor}, is the first to demonstrate stable 2D segmentation at up to $1024{\times}$ compression while maintaining perceptual fidelity at moderate ratios, and first to evaluate synapse and mitochondria detection on highly compressed EM images. We further introduce an ROI-driven workflow that, atop $1024{\times}$ extreme compression, enables selective high-resolution reconstruction of localized regions on demand (e.g., for mitochondria or vesicle analysis). 

\begin{figure*}[t]
  \centering
  \includegraphics[width=1\textwidth]{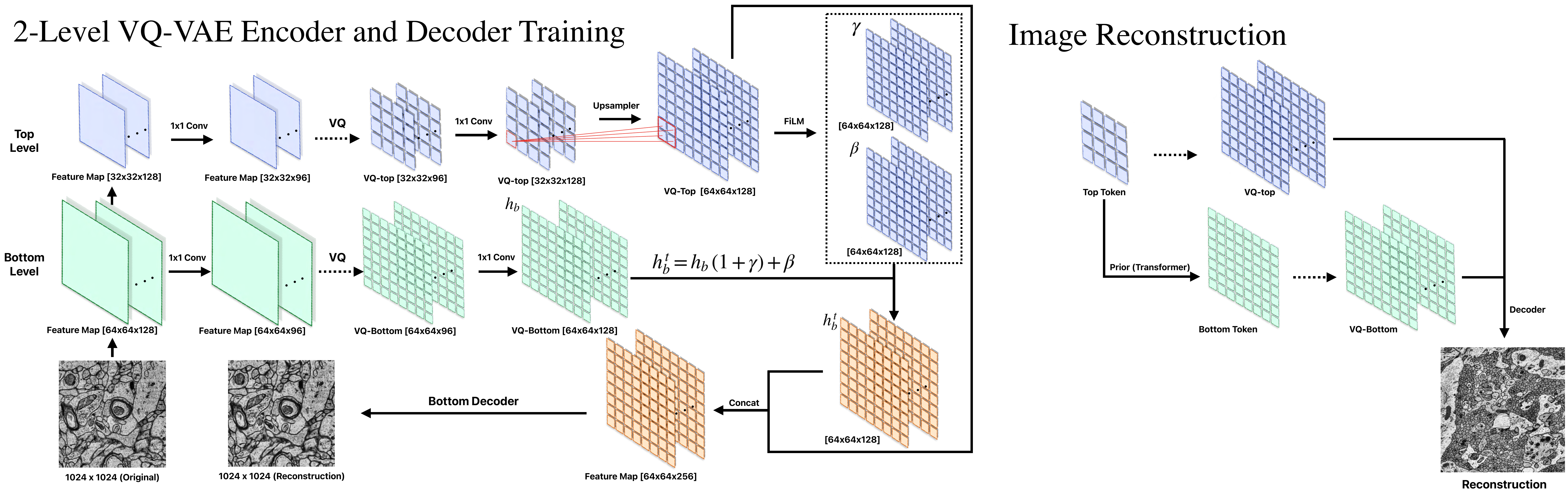}
  \caption{Left: Encoder and Decoder training. Right: Image reconstruction.}
  \label{fig:pipeline}
\end{figure*}

\section{Methodology}
\subsection{Two-Level VQ-VAE Encoder and Decoder Training}
The compression/reconstruction pipeline and the decoder-side fusion used when multi-level latents are present are shown in Fig.~\ref{fig:pipeline}. All experiments use EM sections: H01-human with 3{,}000 images at $1024{\times}1024$ \cite{H01}, and mouse cerebellum with 500 images at $4096{\times}2048$ \cite{Dhanyasi2025DevCerebellum}. Images are converted to single-channel tensors and linearly scaled to $[-0.5,\,0.5]$. Training uses non-overlapping $1024{\times}1024$ tiles. The encoder is a stride-2 convolutional pyramid with $d_s$ downsampling stages, followed by two residual blocks (hidden width 128). A $1{\times}1$ projection yields a 96-D feature per spatial location, vector-quantized with a codebook of $K{=}256$ embeddings; cluster counts and code means are updated by EMA, straight-through estimation passes gradients to the pre-quantized features, and codebook perplexity is monitored. Compression points are set by $d_s\!\in\!\{2,3,4,5\}$, corresponding on $1024^2$ tiles to token grids $256{\times}256$, $128{\times}128$, $64{\times}64$, $32{\times}32$, which yield nominal spatial area reductions of $16{\times}$, $64{\times}$, $256{\times}$, and $1024{\times}$, respectively. In addition, we report an intermediate $128{\times}$ compression by uniformly subsampling the $64{\times}64$ token grid with a checkerboard mask, retaining tokens with $(i{+}j)\bmod 2=0$ and dropping the rest without changing the code dimensionality. The $32{\times}32$/$64{\times}64$ sizes in the figure are illustrative only. The decoder upsamples the quantized feature map to full resolution via $d_s$ stages of transposed convolutions with ReLU and a final $3{\times}3$ prediction head. With both a top and a bottom latent, the upsampled top latent produces channel-wise affine parameters $(\gamma,\beta)$ via $1{\times}1$ convolutions to modulate the bottom quantized feature $h^b$ as $h_b^t = h^b \odot (1{+}\gamma) + \beta$; $h_b^t$ is concatenated with the upsampled top latent and refined by residual blocks before the upsampling stack. The reconstruction loss is
\begin{equation}
\mathcal{L}_\mathrm{rec}=\alpha\|x-\hat{x}\|_{1}
+\beta\bigl(1-\mathrm{MS\text{-}SSIM}(x,\hat{x})\bigr)
+\gamma\|\nabla x-\nabla \hat{x}\|_{1}
\label{eq:lrec}
\end{equation}
\noindent\textit{where} $\mathrm{MS\text{-}SSIM}$ \textit{is multi-scale structural similarity \cite{Wang2003MSSSIM}, and the total objective function is}

\begin{equation}
\mathcal{L}_\mathrm{total}
= \mathcal{L}_\mathrm{rec}
+ \sum_{\ell\in\{\mathrm{top},\mathrm{bot}\}} \lambda_\mathrm{com}^{\ell}\,\mathcal{L}_\mathrm{com}^{\ell},
\label{eq:ltotal}
\end{equation}

where $\mathcal{L}_\mathrm{com}^{\ell}$ is the EMA-based VQ commitment loss for level $\ell$. Unless noted: $\alpha{=}1$, $\beta{=}0.5$, $\gamma{=}0.1$, hidden width $128$, embedding dimension $96$, $K{=}256$, AdamW with a learning rate $2{\times}10^{-4}$ and weight decay $10^{-4}$, batch size $2$, and $100$ epochs. To avoid seams in full-frame reconstruction, we use overlap–add with separable Hann windows: each $1024{\times}1024$ prediction is Hann-weighted, windowed outputs and weights are accumulated at their spatial locations, and the result is normalized by the summed weights. We report PSNR, SSIM, and codebook perplexity on held-out tiles.
\subsection{Image Reconstruction}
We consider two modes with the same decoder. (i) \emph{Top-only direct decoding:} a discrete grid of top tokens (sampled from a prior or taken from an encoder) drives the decoder directly as a single quantized feature, producing compressed images at the ratio governed by the top grid. (ii) \emph{Transformer-augmented two-level reconstruction at the same compression ratio as top-only:} a Transformer prior, conditioned only on the discrete top tokens, predicts the discrete bottom tokens. At inference, given top tokens alone (thus keeping the token budget identical to top-only), the Transformer generates the bottom tokens; the upsampled top latent then modulates the predicted bottom latent via FiLM \cite{Perez2018FiLM}, the two are concatenated, and the decoder outputs the final image. This preserves the compression ratio of top-only while recovering additional texture details.

\begin{figure*}[t] \centering \includegraphics[width=1\linewidth]{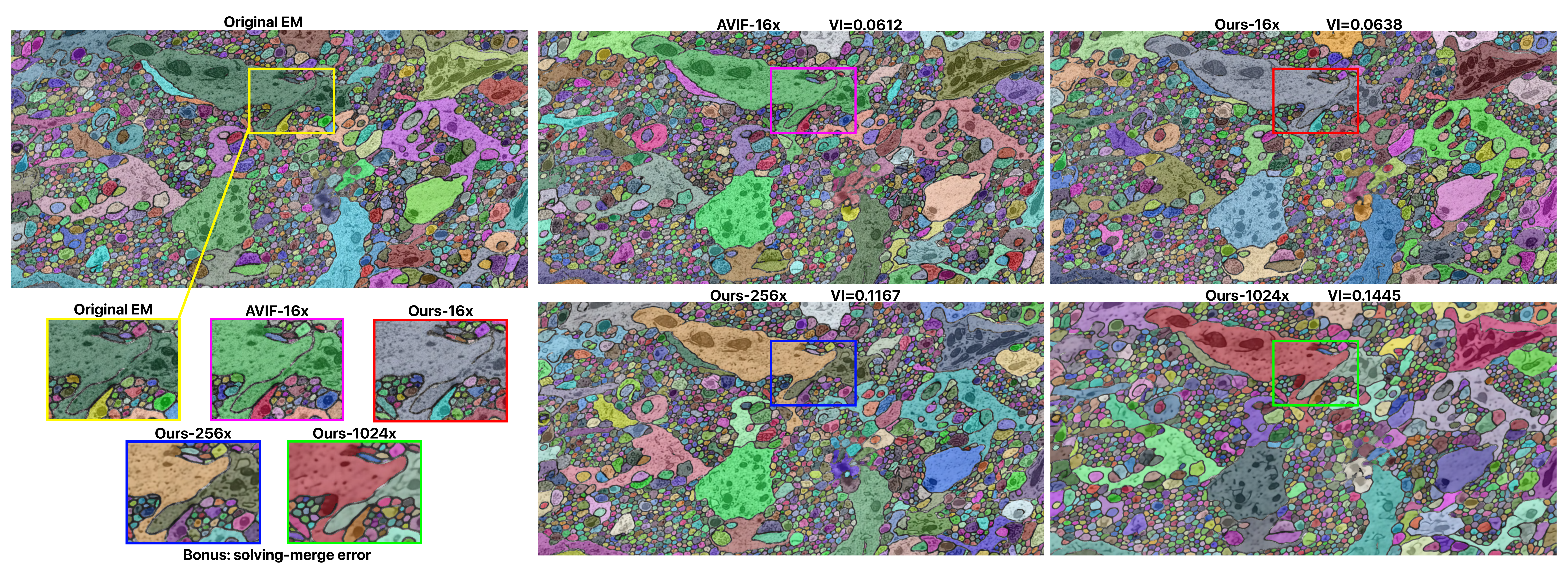} \caption{2D segmentation comparison for AVIF-16x, and Ours-16x, 256x, and 1024x} \label{fig:colorEM} \end{figure*}

\section{Results}
We evaluate compression from two aspects. First, we quantify changes in texture relative to the original EM images across methods and ratios using the structural similarity index measure (SSIM) score. Second, we assess downstream utility by measuring the accuracy of machine-learning performance at different compression ratios.

\begin{figure}[t] \centering \includegraphics[width=1\linewidth]{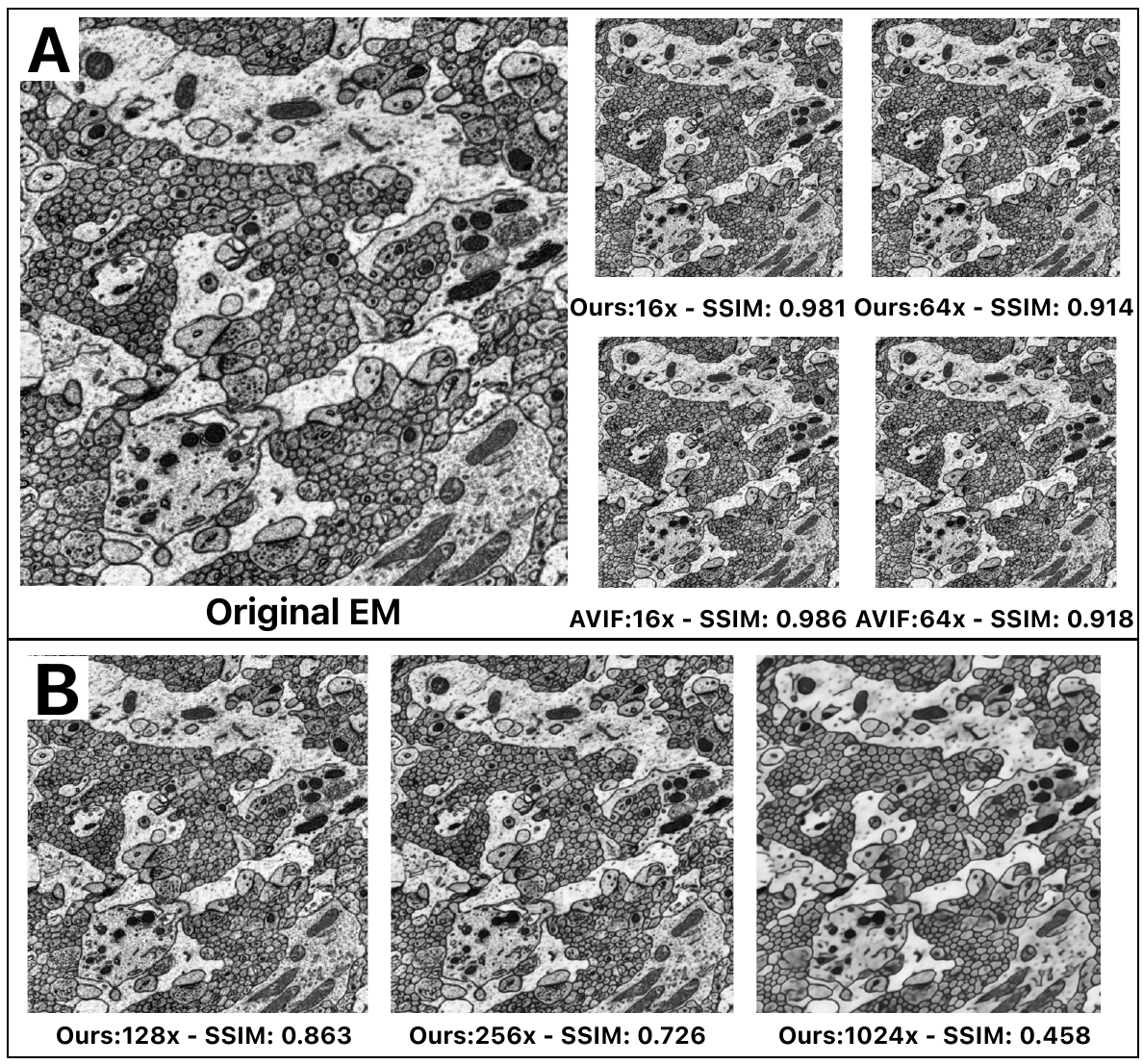} \caption{SSIM comparison. (A) Ours vs.\ AVIF. (B) Extended ratios of ours.} \label{fig:conpressedEM} \end{figure}

We benchmark against AVIF. Because AVIF becomes impractical at $\geq 128{\times}$ under our quality/bitrate criteria, we run direct comparisons at $16{\times}$ and $64{\times}$. For higher compression ($\geq 128{\times}$: $128{\times}$, $256{\times}$, $1024{\times}$), AVIF is out of range, so we report only our method. From the quantitative results (Tab.~\ref{tab:ssim}) and visual examples (Fig.~\ref{fig:conpressedEM}), while AVIF achieves a slightly higher SSIM at $16{\times}$ and $64{\times}$, our method is highly competitive (within $0.005$) and allows for much higher compression ratios, at $128{\times}$ and $256{\times}$, SSIM remains above $0.72$, and at $1024{\times}$ it remains above $0.45$.

\begin{table}[t]
  \centering
  \caption{SSIM  vs.\ compression ratio. Values are dataset-wide averages over 500 EM images on H01 and 100 on mouse cerebellum (test datasets).}
  \label{tab:ssim}
  \setlength{\tabcolsep}{4pt}
  \renewcommand{\arraystretch}{1.12}
  \footnotesize
  \resizebox{\columnwidth}{!}{%
  \begin{tabular}{llccccc}
    \hline
    Dataset & Method & $16{\times}$ & $64{\times}$ & $128{\times}$ & $256{\times}$ & $1024{\times}$ \\
    \hline
    H01-human         & Ours & 0.982 & 0.914 & 0.862 & 0.728 & 0.459 \\
                      & AVIF & 0.986 & 0.916 &  --   &  --   &  --   \\
    \hline
    Mouse cerebellum  & Ours & 0.979 & 0.908 & 0.855 & 0.715 & 0.445 \\
                      & AVIF & 0.984 & 0.911 &  --   &  --   &  --   \\
    \hline
  \end{tabular}}
\end{table}

\begin{figure}[H] \centering \includegraphics[width=1\linewidth]{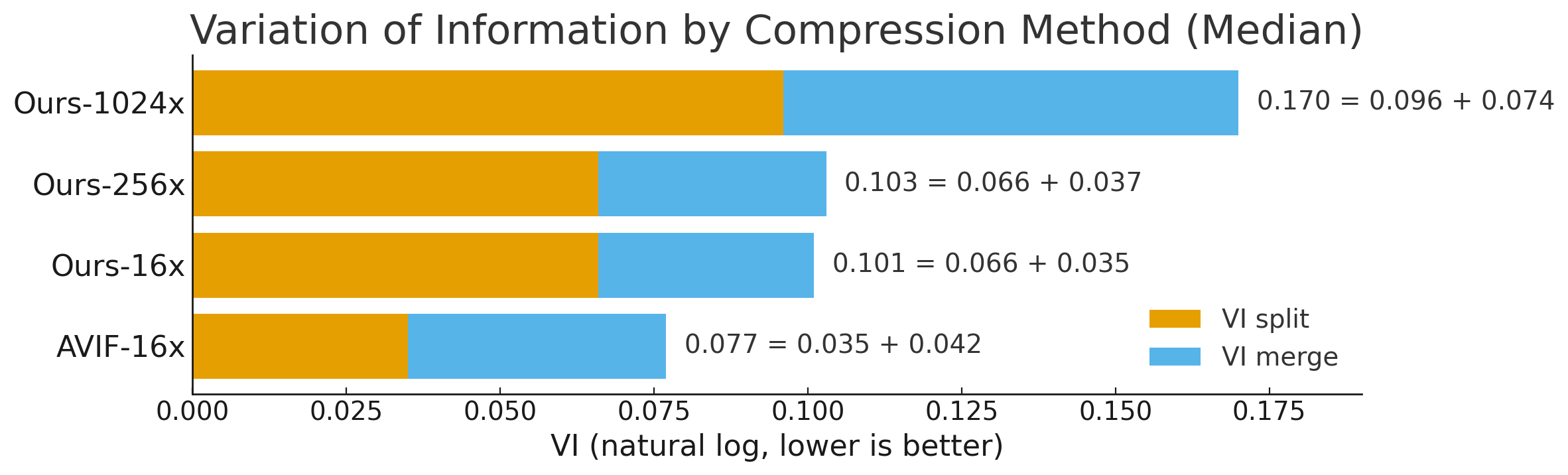} \caption{Median Variation of Information (lower is better). Bars decompose VI into split and merge.} \label{fig:seg_results} \end{figure}

To evaluate 2D segmentation performance, we use a mouse cerebellum EM dataset \cite{Dhanyasi2025DevCerebellum}. Pseudo--ground-truth labels are produced by a strong model \cite{pavarino2023membrain} and we train and evaluate on compressed images, demonstrating transfer learning from uncompressed to compressed domains. As shown in Fig.~\ref{fig:seg_results}, our median VI at \(16\times\) is comparable to AVIF (total VI \(<0.03\)). Increasing the compression from \(16\times\) to \(256\times\) keeps the VI essentially unchanged (\(\Delta\)VI \(<0.002\)). A qualitative example in Fig.~\ref{fig:colorEM} shows that the quality of 2D segmentation is well preserved in compressed images. As a bonus, in some cases (Fig.~\ref{fig:colorEM}, bottom-left) our generative reconstructor inpaints faint or broken membranes (existing in the original EM) and thereby reduces merge errors, especially at lower compression ratios.

Synapse prediction remains stable even under $512{\times}$ compression: the accuracy differs from using the original images by less than $0.005$ (Tab.~\ref{tab:synapse_h01}). We use the same state-of-the-art synapse detector, a 3D U-Net trained with the Budgeted Broadcast Learning Rule \cite{Meirovitch2025BudgetedBroadcast} under an identical training set on the SmartEM project \cite{Meirovitch2025SmartEM}; the only difference is whether the input stack is compressed by $512{\times}$ prior to training.

\begin{table}[t]
  \centering
  \caption{Synapse prediction on H01-human.}
  \label{tab:synapse_h01}
  \setlength{\tabcolsep}{6pt}
  \renewcommand{\arraystretch}{1.15}
  \footnotesize
  \resizebox{\columnwidth}{!}{%
  \begin{tabular}{lcc}
    \hline
    Input (3 seeds) & Mean (\%)  & $\Delta$ vs.\ original (\%) \\
    \hline
    Original EM        & 94.1 ($\pm$ 0.2) & -- \\
    $512{\times}$ compressed EM & 93.9 ($\pm$ 0.3) & $-0.2$ \\
    \hline
  \end{tabular}}
\end{table}

\begin{table}[t]
  \centering
  \caption{Cross-dataset SSIM at 16$\times$ compression.}
  \label{tab:transfer}
  \setlength{\tabcolsep}{6pt}
  \renewcommand{\arraystretch}{1.15}
  \footnotesize
  \resizebox{\columnwidth}{!}{%
  \begin{tabular}{lcc}
    \hline
    Test (SSIM) & H01-human & Mouse cerebellum \\
    \hline
    H01-human       & 0.982 & 0.976 \\
    Mouse cerebellum & 0.968 & 0.979 \\
    \hline
  \end{tabular}}
\end{table}

\begin{figure}[H]
  \centering
  \includegraphics[width=1\linewidth]{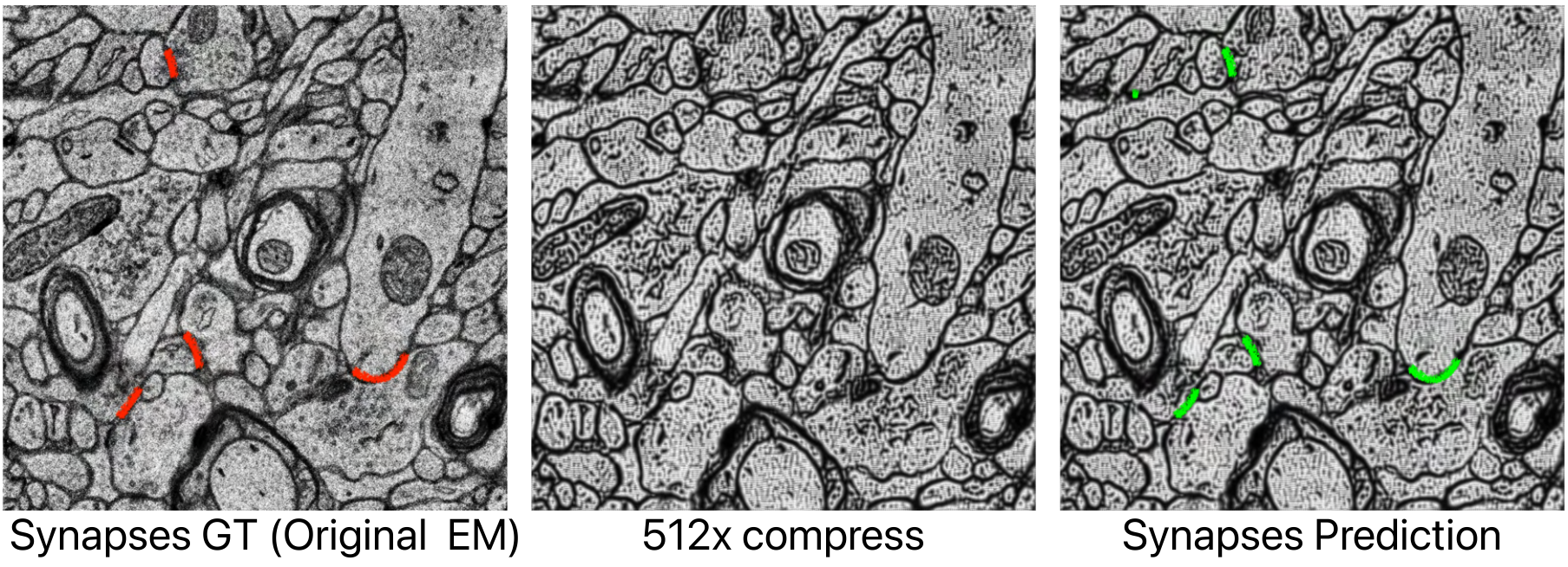}
  \caption{Synapse prediction on the compressed EM}
  \label{fig:synapses}
\end{figure}

\begin{figure}[t]
  \centering
  \includegraphics[width=1\linewidth]{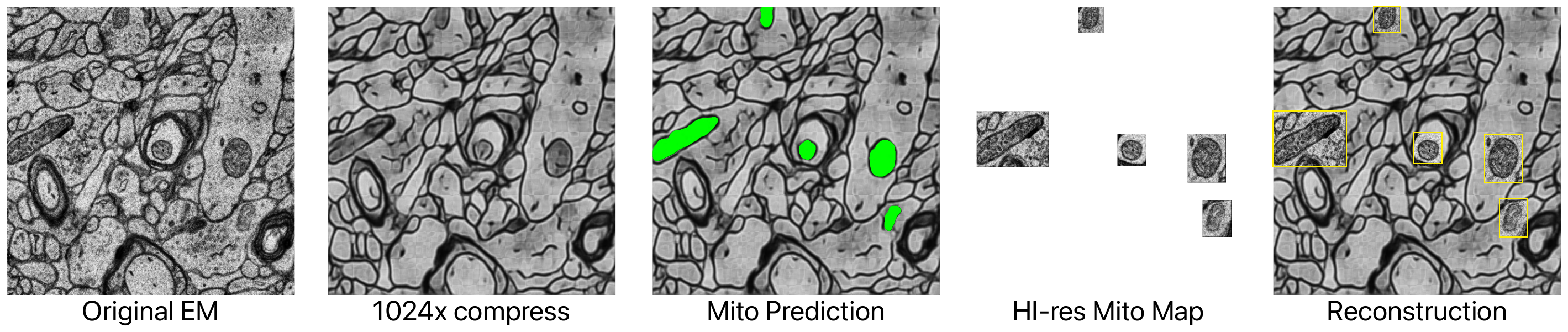}
\caption{Selective high-resolution mitochondria from 1024$\times$ compressed EM.}
  \label{fig:mito}
\end{figure}

For mitochondria prediction, performance remains strong even at $1024{\times}$ compression: object size is largely preserved, though internal texture degrades at ultra-high ratios. To mitigate this, we introduce a selective high-resolution EM pipeline operating from $1024{\times}$-compressed latents (Fig.~\ref{fig:mito}): pretrained detectors (e.g., mitochondria or vesicle networks) first localize targets on the compressed input; based on these predictions, we crop the corresponding regions directly from the uncompressed image and either apply mild AVIF compression or store them as PNGs. These sub-regions are then concatenated with (i.e., stored alongside) the global $1024{\times}$ representation on disk.

\section{Discussion and Future Work}
\noindent Taken together, our results suggest reframing EM compression as a token-level interface between storage and analysis that enables \emph{pay-as-you-decode}: extreme ratios for bulk storage and fast screening, with selective high-resolution decoding only where biological detail matters. In Table~\ref{tab:transfer}, our results show that a model trained on one dataset can perform well on another, even without multi-dataset training, suggesting that this architecture may serve as a backbone for the future foundation model for EM compression across connectomic datasets. Looking forward, training can be augmented with lightweight ``detail experts'' for vesicles, synapses, mitochondria, and membranes; via cross-attention, these heads can modulate FiLM or intermediate latents to enrich fine texture without increasing the token budget. On the image reconstruction side, a small adapter that maps discrete top tokens directly to decoder-ready features would bypass explicit VQ-top dequantization/embedding lookup, further reducing latency.

\section{Acknowledgments}
We thank Nagaraju Dhanyasi for providing the mouse cerebellum datasets, and the H01 project for making the human cerebral cortex datasets publicly available.

\bibliographystyle{IEEEbib}

\end{document}